\documentclass[preprint,review,12pt]{elsarticle}

\usepackage{amssymb}
\usepackage{amsmath}
\usepackage{url}
\usepackage{booktabs}
\usepackage{multirow}
\usepackage{multicol}
\usepackage{makecell}
\usepackage{caption}
\usepackage{tabularray}
\usepackage{setspace,tabularx}
\usepackage{bbding}
\usepackage[dvipsnames]{xcolor}
\DeclareCaptionFont{helv}{\sf\small}
\let\oldtabularx\tabularx
\let\endoldtabularx\endtabularx
\renewenvironment{tabularx}[2]
  {\singlespacing
   \noindent
   \oldtabularx{#1}{#2}}
  {\endoldtabularx}

\usepackage{lineno}

\journal{Knowledge-Based Systems}

\begin{document}

\begin{frontmatter}



\title{EntGPT: Entity Linking with Generative Large Language Models}


\author[label1,label3]{Yifan Ding}
\affiliation[label1]{organization={University of Notre Dame},
city={Notre Dame},
state={Indiana},
country={USA}}

\affiliation[label2]{%
organization={Deloitte \& Touche LLP},
city={New York},
state={New York},
country={USA},
}
\affiliation[label3]{Both authors contributed equally to this research.}
\affiliation[label4]{Corresponding author: tweninger@nd.edu}

\author[label1,label3]{Amrit Poudel}
\author[label1]{Qingkai Zeng}
\author[label1,label4]{Tim Weninger}

\author[label2]{Balaji Veeramani}

\author[label2]{Sanmitra Bhattacharya}

\begin{abstract}

Entity Linking in natural language processing seeks to match text entities to their corresponding entries in a dictionary or knowledge base. Traditional approaches rely on contextual models, which can be complex, hard to train, and have limited transferability across different domains. Generative large language models like GPT offer a promising alternative but often underperform with naive prompts. In this study, we introduce EntGPT, employing advanced prompt engineering to enhance EL tasks. Our three-step hard-prompting method (EntGPT-P) significantly boosts the micro-$F_1$ score by up to 36\% over vanilla prompts, achieving competitive performance across 10 datasets without supervised fine-tuning. Additionally, our instruction tuning method (EntGPT-I) improves micro-$F_1$ scores by 2.1\% on average in supervised EL tasks and outperforms several baseline models in six Question Answering tasks. Our methods are compatible with both open-source and proprietary LLMs. All data and code are available on GitHub at \url{https://github.com/yifding/In_Context_EL}.

\end{abstract}




\end{frontmatter}





\section{Introduction}

The advent of Large Language Models (LLMs) has generated significant excitement and concern in the fields of natural language processing (NLP) and artificial intelligence (AI). While their text generation and unstructured reasoning capabilities are impressive~\cite{Raffel-JMLR'20-T5, Radford-OpenAI_Blog'18-GPT, Radford-OpenAI_Blog'19-GPT2, Brown-NIPS'20-GPT3, Touvron-arXiv'23-Llama2}, their ability to produce structured output remains underdeveloped and relatively unexplored~\cite{Zhu-AAAI'22-JAKET, Dong-arXiv'23-Head-to-Tail}.

Entity disambiguation (ED), a crucial task in information extraction (IE), involves linking text fragments representing real-world entities to entries in structured knowledge bases like encyclopedias or dictionaries. These links can enhance various downstream processes by leveraging existing data and relationships within the knowledge base. Integrating deep neural network models, such as Transformers~\cite{Vaswani-NIPS'17-attention}, with the symbolic reasoning capabilities of knowledge bases has been termed the ``Third Wave of AI''\cite{Darpa-Darpa'18-Next_AI}. This convergence is essential for advancing problem-solving and reasoning in AI systems\cite{Brooks-AI'81-Symbolic, Hitzler-Book'22-Neuro}. For this vision to materialize, it is critical that LLMs are effectively linked to the symbolic entities they reference.

Several models predict entity labels from Wikipedia links~\cite{Ganea-EMNLP'17-deep_ed,Kolitsas-CONLL'18-end2end}, Web hyperlinks~\cite{Wu-EMNLP'20-BLINK}, or info-boxes~\cite{Ayoola-NAACL'22-ReFinED, Bhargav-NAACL'22-Dbpedia_Entity_Type}, considering symbolic consistency among entities~\cite{Hu-KBS'20-Graph_Entity} and between entities and entity types~\cite{Raiman-AAAI'18-deeptype, Tedeschi-EMNLP'21-NER4EL, Raiman-AAAI'22-deeptype2,Phan-TKDE'19-Pair_link, Xiao-EMNLP'23-CoherentED}. These methods have been successful in closed-world scenarios where data is clean and complete~\cite{Hoffart-EMNLP'11-CONLL03_entity}, and in open-world scenarios where data may be incomplete but well-described~\cite{Logeswaran-ACL'19-zero, Wu-EMNLP'20-BLINK}.

Entity disambiguation tasks have traditionally been framed as entity classification~\cite{Ganea-EMNLP'17-deep_ed}. More recent approaches have included machine reading comprehension~\cite{Barba-ACL'22-Extend}, dense retrieval~\cite{Wu-EMNLP'20-BLINK}, question answering~\cite{Zhang-ICLR'22-Entqa}, sequence-to-sequence generation~\cite{DeCao-ICLR'21-GENDRE}, and, most recently, prompt engineering with LLMs~\cite{Xiao-EMNLP'23-Instruction}.


The advantage of prompt-based methods lies in their ability to query a language model for additional knowledge and context understanding, significantly enhancing performance in zero-shot and few-shot settings~\cite{Li-arXiv'23-GPT-NER, Dong-arXiv'23-Head-to-Tail}. Existing prompt-based methods for information extraction can be categorized into three main paradigms: type-oriented, span-oriented, and generation. Type-oriented tasks~\cite{Ding-ACL'22-Prompt_Entity_Typing} focus on locating the appropriate class for a given mention within the original documents. Span-oriented methods~\cite{Cui-ACL'21-template_NER_BART} enumerate possible spans and assign corresponding class labels. Generation methods aim to transform information extraction into a sequence generation schema~\cite{Li-arXiv'23-GPT-NER}. Despite their promise, the use of LLMs for information extraction tasks~\cite{Qin-arXiv'23-Chatgpt_NLP_Task, Gonzalez-arXiv'23-ChatGPT_Historical_Entities, wei2023zero, li2023prompting} typically focuses on label sets with only a few dozen entity types, far fewer than the 6 million entities required for comprehensive ED.

The goal of the present work is to address common failure cases in ED. Most errors in ED stem from rare entities, but auxiliary knowledge from LLMs can help mitigate these challenging cases. Additionally, most ED models require a ranking step to select the best entity from several candidates within a given context. Generative LLMs, with their exceptional classification abilities~\cite{xu2023large, fei2022lasuie, josifoski2022genie}, can naturally integrate into the entity ranking step, leveraging their extensive background knowledge.

\begin{figure}[t]
    \centering
    \includegraphics[width=\linewidth]{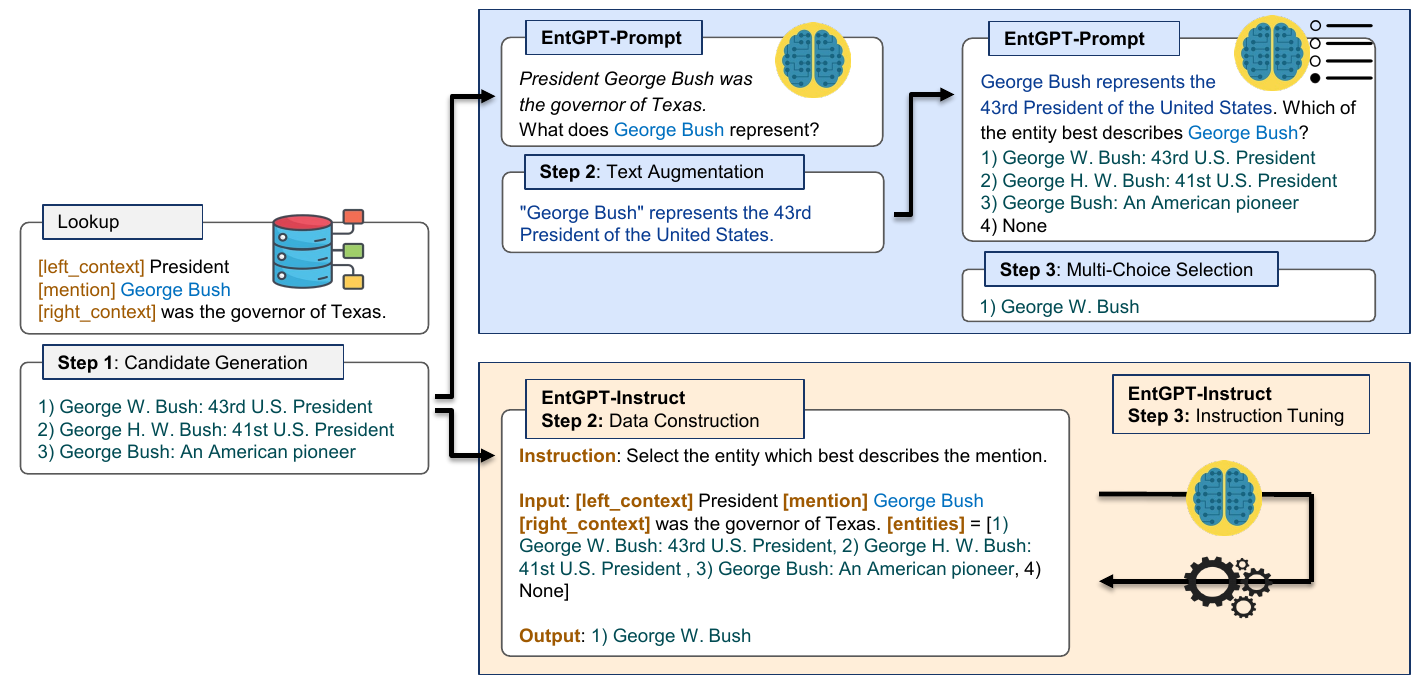}
    \vspace{-0.8cm}
    \caption{Pipeline of EntGPT frameworks: Given input document with the annotated mention, both of the approaches (1) generate relevant entity candidates. EntGPT-Prompt then (2) finds an auxiliary content of the annotated mention and then (3) selects the corresponding entity from the list. EntGPT-Instruct (2) constructs an instruction dataset, and (3) is instruction-tuned on the curated data.}
    \label{fig:model}
\end{figure}


We explore two distinct approaches for tackling the ED task with LLMs, as illustrated in Fig.~\ref{fig:model}. The first approach is a multi-step prompting framework called EntGPT-Prompting (EntGPT-P). This method first generates a set of entity candidates for mentions in a document. Then, it uses the LLM to generate auxiliary content, aiding the final selection of the correct entity from the candidate set. Results across various tasks and datasets show that EntGPT-P achieves performance comparable to supervised models without requiring training or fine-tuning on human-annotated data. The second approach, EntGPT-Instruction Tuning (EntGPT-I), involves further training on instructions (prompt-response pairs)~\cite{zhang2023instruction}. EntGPT-I demonstrates more controllable and predictable behavior, significantly outperforming EntGPT-P in micro-$F_1$ scores across ED benchmarks. Additionally, EntGPT-I was evaluated on six closed-world question-answering tasks, achieving state-of-the-art (SOTA) performance in the zero-shot setting and significantly improving accuracy scores over leading commercial LLMs.

Throughout the remainder of this paper, we use the following abbreviations: GPT3.5 for gpt-3.5-turbo, Llama for Llama-65B, Llama2 for Llama2-70B, and Llama2-chat for Llama2-70B-chat unless stated otherwise.

\section{EntGPT}

Here, we formally define the ED task and describe the general pipeline employed in both the prompt-based and instruction-tuning approaches to ground generative LLMs with entities from a knowledge base.

Given a document represented as a sequence of tokens $\mathcal{D}$ and a set of subsequences $M = \{m_1, \cdots, m_n\}$ (\textit{i.e.}, mentions) contained in $\mathcal{D}$. The goal of the ED task is to establish a mapping for each mention $m_i \in M$ to its corresponding entity $e$ in the set $\mathcal{E}$ representing entities in some knowledge base (such as Wikipedia or DBpedia).

\subsection{EntGPT-P: Entity Disambiguation with Multi-Step Prompting}

The task of grounding generative LLMs to knowledge bases is a natural use case of ED, where text sequences are matched with relevant knowledge base entries for downstream tasks.

EntGPT-Prompt (EntGPT-P) formulates the ED task in three steps: (1) Generate and filter entity candidates from the knowledge base $\mathcal{E}$. (2) Augment each mention by extracting relevant information with a prompt for an LLM. (3) Combine candidates from Step 1 with context from Step 2, forming a multi-choice question prompt for an LLM.

\paragraph{Step 1: Entity Candidate Generation} For each mention $m$ in document $D$, we generate a subset of entity candidates $\mathcal{E}_c$ from the knowledge base $\mathcal{E}$. We use two strategies: (1) A probabilistic approach (\textit{Prior}) based on statistical information of hyperlinks, surfacing syntactically similar candidates $\mathcal{E}_p$~\cite{Ganea-EMNLP'17-deep_ed}. (2) A dense retrieval model (BLINK) that uses cleaned Wikipedia hyperlinks to generate additional candidates $\mathcal{E}_r$~\cite{Wu-EMNLP'20-BLINK}. The final candidate set is $\mathcal{E}_c = \mathcal{E}_p \cup \mathcal{E}_r$, from which we select the top-10 candidates.

\paragraph{Step 2: Augmentation by Prompting} To distinguish between syntactically similar candidates, we augment mentions with relevant information by asking the LLM, ``What does {\texttt{mention}} represent?'' For example, asking about ``George Bush'' generates auxiliary content $\mathcal{A}$ that considers the document's context and the LLM's world knowledge.

\paragraph{Step 3: Multiple-choice Selection by Prompting} Using the entity candidates $\mathcal{E}_c$ from Step 1 and auxiliary content $\mathcal{A}$ from Step 2, we prompt the LLM to select the most appropriate entity $e \in \mathcal{E}_c$. Each candidate's first sentence from the knowledge base is included as a reference. To permit cases where the entity is not in $\mathcal{E}_c$, we include a None option.

\subsection{EntGPT-I: Entity Disambiguation with Instruction Tuning}
This section outlines the construction of the EntGPT-I model, including the creation of an instruction dataset and the instruction-tuning process.

\paragraph{Instruction Dataset Construction} An instruction dataset typically includes an instruction, optional input for additional context, and the expected output~\cite{ji2023survey}. In our approach, we transform the AIDA dataset~\cite{hoffart2011robust}, a large manually annotated dataset for entity linking and disambiguation, into the required format (instruction, input, output).

\paragraph{Instruction Tuning} We conduct supervised fine-tuning of two generative models, GPT3.5 and Llama2-chat, using the curated instruction dataset. GPT3.5 is fine-tuned using the OpenAI API platform~\cite{kublik2023gpt}, while Llama2-chat is fine-tuned using Replicate API\footnote{https://replicate.com/docs/get-started/python}. Fine-tuning involves one epoch with default hyperparameters. Following fine-tuning, EntGPT-I is evaluated on both ED and question answering tasks.

\section{Methodology}

We evaluated these models using ten public entity disambiguation benchmark datasets and six widely recognized question answering datasets. The entity disambiguation task involved linking dataset mentions to their corresponding entities in a knowledge base, while the question answering task required selecting the correct answer from multiple choices.

\paragraph{Datasets}

Following established practices in entity disambiguation (ED) research~\cite{Guo-Semantic_Web'16-Random_Walk, Kolitsas-CONLL'18-end2end, Ayoola-NAACL'22-ReFinED, Barba-ACL'22-Extend}, we tested our models on a total of ten ED datasets: AIDA~\cite{Milne-CIKM'08-Learning_to_Link}, MSNBC~\cite{Cucerzan-EMNLP'07-MSNBC}, Aquaint (AQUAINT/AQU)\cite{Milne-CIKM'08-Learning_to_Link}, ACE2004~\cite{Ratinov-ACL'11-Local_Global_Disambiguation}, WNED-WIKI~\cite{Guo-Semantic_Web'16-Random_Walk}, WNED-CWEB~\cite{Gabrilovich-Dataset'13-Clueweb}, KORE50~\cite{Hoffart-CIKM'12-KORE}, OKE-2015~\cite{Nuzzolese-SWEC'15-OKE15}, OKE-2016\cite{Nuzzolese-SWEC'16-OKE16}, REU~\cite{Roder-LREC'14-n3}, and RSS~\cite{Roder-LREC'14-n3}. Each experiment utilized Wikipedia as the knowledge base. Following prior work~\cite{Guo-Semantic_Web'16-Random_Walk}, we preprocessed datasets by removing spurious mentions not found in the knowledge base, as well as duplicate and empty documents.

For ED, we categorize the comparison models into  three different groups according to their fine-tuning data: Wikipedia and AIDA (Wiki+AIDA), AIDA-only (AIDA), and no fine-tuning (zero-shot). We compared the performance of our models against seven recent SOTA models. In the question answering task we categorize the baselines into two different groups: supervised and zero-shot. In the supervised setting, we follow common experimental methods to report the performance against three recent SOTA models.  In the zero-shot setting, we use several LLM foundation models. 

Additionally, we evaluated EntGPT-I (GPT3.5) on six commonsense reasoning benchmarks: ARC-C~\cite{clark2018think}, OBQA~\cite{mihaylov2018can}, aNLI~\cite{bhagavatula2019abductive}, CSQA~\cite{talmor2018commonsenseqa}, PIQA~\cite{bisk2020piqa}, and SIQA~\cite{sap2019socialiqa}.

%

\subsection{Evaluation Metrics}
To maintain a fair comparison across datasets, we use the \textit{in-KB} micro-F1 score as our evaluation metric for ED following the example of Guo and Barbosa~\cite{Guo-Semantic_Web'16-Random_Walk}. Specifically, being in-KB requires that ground truth mentions correspond to existing knowledge base entries. Empty or out-of-KB entities are removed in the evaluation process. The mean Micro-F1 score is taken per-mention. Although a model may predict non-entities, each mention will have some corresponding ground truth entity. For question answering, we use accuracy as the evaluation metric. 

\section{Results}


\subsection{Entity Disambiguation}

\begin{table}[t]
\centering
\footnotesize
\caption{Test micro-F1 scores on standard six benchmarks. With each setting, best scores are highlighted in bold. \textdagger{} indicates that the results are obtained directly from the source paper.}
\begin{tabular}{l  p{3.9cm} | llllll|l}
\toprule
& {\textbf{Model}}
& \textbf{AIDA} &  \textbf{MSN} &
\textbf{AQU} & \textbf{ACE04} & \textbf{CWEB} & \textbf{WIKI} & \textbf{avg} \\
\midrule
\multirow{6}{*}{\rotatebox[origin=c]{90}{Wiki+AIDA}}
& \textbf{REL}\textsuperscript{\textdagger}~\cite{Hulst-SIGIR'20-REL} & 0.928 & 0.935 & 0.873 & 0.897 & 0.776  & 0.780 & 0.864 \\
& \textbf{End2End}\textsuperscript{\textdagger}~\cite{Kolitsas-CONLL'18-end2end}&0.891  & 0.933 & 0.894 & 0.892 &0.760 &0.740 &0.851 \\
& \textbf{GENRE}\textsuperscript{\textdagger}~\cite{DeCao-ICLR'21-GENDRE} & 0.933 & 0.943 & 0.899 & 0.901 & 0.773 & 0.874 & 0.887 \\
& \textbf{ReFinED}\textsuperscript{\textdagger}~\cite{Ayoola-NAACL'22-ReFinED} &\textbf{0.939} & 0.941 & 0.918 & 0.908 & 0.794 & 0.874 & 0.894\\
& \textbf{ExtEnD}\textsuperscript{\textdagger}~\cite{Barba-ACL'22-Extend} &0.926 &0.947 &0.916 &0.918 &0.777 &0.888 & 0.895\\
& \textbf{CoherentED}\textsuperscript{\textdagger}~\cite{Xiao-EMNLP'23-CoherentED} & 0.894 & \textbf{0.963} & \textbf{0.946} & \textbf{0.934}& \textbf{0.811} & \textbf{0.906} & \textbf{0.909}\\

\midrule
\multirow{5}{*}{\rotatebox[origin=c]{90}{AIDA}}
&\textbf{GENRE}\textsuperscript{\textdagger}~\cite{DeCao-ICLR'21-GENDRE} &0.886	&0.881	&0.771 &0.823 &0.719  & 0.717 &0.795 \\
&\textbf{ner4el}\textsuperscript{\textdagger}~\cite{Tedeschi-EMNLP'21-NER4EL} & \textbf{0.925}	&0.892	&0.695 &0.913 &0.685  & 0.640 &0.792 \\
&\textbf{ExtEnd}\textsuperscript{\textdagger}~\cite{Barba-ACL'22-Extend} & 0.900	&\textbf{0.945}	&0.879 &0.889 &0.766  & 0.767 &0.858\\
&\textbf{EntGPT-I (Llama2)}    & 0.691 & 0.744 & 0.583 & 0.681 & 0.556 & 0.604 & 0.643 \\
&\textbf{EntGPT-I (GPT3.5)} & 0.920	&0.922	&\textbf{0.906} &\textbf{0.937} &\textbf{0.795}  &\textbf{0.791} &\textbf{0.879}\\

\midrule
\multirow{4}{*}{\rotatebox[origin=c]{90}{zero-shot}}
& \textbf{Llama2} &0.432 &0.401 &0.441 & 0.381 & 0.347  & 0.451 & 0.408\\
& \textbf{EntGPT-P (Llama2) } & 0.708 & 0.741 & 0.635 & 0.746 & 0.584 & 0.632 & 0.674 \\
& \textbf{GPT3.5} &0.746  &0.821 & 0.753 & 0.853 & 0.638 & 0.684 & 0.749\\
& \textbf{EntGPT-P (GPT3.5) } & \textbf{0.821} 	&\textbf{0.867}	&\textbf{0.791} &\textbf{0.918}	&\textbf{0.709}  & \textbf{0.808} & \textbf{0.819} \\


\bottomrule
\end{tabular}

\label{tab:main_experiment}
\end{table}

A performance comparison over six entity disambiguation benchmarks is presented in Table~\ref{tab:main_experiment}. We categorized these experiments into three groups based on the data used during the fine-tuning process. Compared to the available model categories, EntGPT-P follows the zero-shot setting and EntGPT-I uses instruction tuning (on AIDA only) and can be considered as a supervised model. 

From fine-tuning only on the AIDA dataset, we observed that EntGPT-I achieves SOTA micro-F1  performance on AQUAINT, ACE2004, CWEB, and WIKI. Moreover, EntGPT-I achieved an overall mean improvement of +2.1\% micro-\textit{$F_1$} score across all six datasets. At the same time, it also achieved comparable performance against the CoherentED model, which was fine-tuned on data from Wiki and AIDA and outperforms other models. 

We also find that EntGPT-P outperforms the other models across six different ED benchmarks in the zero-shot setting. EntGPT-P reports an average micro-{$F_1$} improvement of +7.0\% over the second-best model. 

\begin{table}[t]
\centering
\footnotesize
\caption{Test micro-$F_1$ scores on five ED benchmarks. We compare EntGPT-P and EntGPT-I with other open-sourced ED methods. Highest micro-$F_1$ scores are highlighted in bold, second highest micro-$F_1$ scores are underlined. The experiments are re-computed from source code to compare entity names for evaluation.}
\begin{tabular}{p{4.0cm} | lllll|l}
\toprule
{\textbf{Model}}
& \textbf{KORE} & \textbf{OKE15} & \textbf{OKE16} &
\textbf{REU} & \textbf{RSS} & \textbf{avg} \\

\midrule
\textbf{REL} & 0.618 & 0.705 & 0.749 & 0.662 & 0.680 &0.682 \\
\textbf{End2End} & 0.569& 0.767 & 0.783& 0.677& 0.720&0.703 \\

\textbf{GENRE} &0.542 &0.640 &0.708 & 0.697 & 0.708 &0.659\\
\textbf{ReFinED} &0.567 & \underline{0.781} & \underline{0.794}	& 0.680 &0.708	&0.706 \\

\textbf{GPT3.5} & 0.707 & 0.696 & 0.687 & 0.688 & 0.767 & 0.709 \\

\midrule
\textbf{EntGPT-P (GPT3.5)} & \underline{0.716}	&0.767 & 0.770	& \underline{0.785} & \underline{0.808}	&\underline{0.769} \\
\textbf{EntGPT-I (GPT3.5)} & \textbf{0.753} & \textbf{0.825}  & \textbf{0.819}	& \textbf{0.808} & \textbf{0.825}	&\textbf{0.806} \\

\bottomrule
\end{tabular}   

\label{tab:kore50}
\end{table}

We further evaluated our approach on five entity benchmarks from General Entity Annotator Benchmarking Framework (GERBIL)~\cite{Usbeck-WWW'15'-GERBIL}. Because previous work does not evaluate on these datasets, we could only consider open-source entity disambiguation models for comparison. We found that both EntGPT-I and EntGPT-P achieved improvement over the other SOTA models as shown in Table~\ref{tab:kore50}. 

\begin{table}[t]
    \centering
    \footnotesize
    \caption{Accuracy on six QA datasets. Most accurate zero-shot scores are highlighted in bold, second most accurate zero-shot scores are underlined. Supervised learning methods are fine-tuned on each dataset. \textdagger{} indicates that the results are obtained directly from the source paper.}
\begin{tabular}{l p{4.0cm}|llll lll |l}

\toprule

& {\textbf{Model}} 
& \textbf{ARC-C} & \textbf{OBQA} & \textbf{ANLI} &
\textbf{CSQA} & \textbf{PIQA} & \textbf{SiQA} \\
\midrule
\multirow{4}{*}{\rotatebox[]{90}{supervised}} 
& \textbf{ROBERTa}\textsuperscript{\textdagger}~\cite{liu2019roberta} &0.430	&0.649	 &0.827	& 0.805	&0.794	&0.759\\
& \textbf{QAGNN}\textsuperscript{\textdagger}~\cite{yasunaga2021qa} &0.444	&0.678	 &0.830	& 0.734	&0.796	&0.757\\
& \textbf{GreaseLM}\textsuperscript{\textdagger}~\cite{zhang2022greaselm} &0.447	&0.669	 &0.833	& 0.742	&0.796	&0.833\\
& \textbf{DRAGON}\textsuperscript{\textdagger}~\cite{yasunaga2022deep} &0.486	&0.720	 &0.840	& 0.760	&0.811	&0.768\\
\midrule
\multirow{6}{*}{\rotatebox[]{90}{zero-shot}} 
& \textbf{LLAMA-65B}\textsuperscript{\textdagger}~\cite{touvron2023llama} &{0.560}	&0.602	&-	&-	&\underline{0.828}	&0.523\\
& \textbf{FLAN-137B}\textsuperscript{\textdagger}~\cite{wei2021finetuned} &0.631	&\underline{0.784}	&-	&-	&0.805	&-\\
& \textbf{Llama-2-70B}\textsuperscript{\textdagger}~\cite{Touvron-arXiv'23-Llama2} &0.574 &0.602 &- &- &\underline{0.828} &0.507 \\ 
& \textbf{Llama-2-70B-chat} &0.613 &0.648 &\underline{0.720} &0.590 &0.670 &0.689 \\ 
& \textbf{GPT3.5} &\underline{0.822}	&0.772	&0.710	&\underline{0.654}	&0.801	&\underline{0.695}\\
& \textbf{EntGPT-I (GPT3.5)} &\textbf{0.840}	&\textbf{0.804} &\textbf{0.754}	&\textbf{0.700} &\textbf{0.830}	&\textbf{0.706} \\

\bottomrule
\end{tabular}
    \label{tab:qa}
\end{table}

\subsection{Question Answering}
We also extended our study to the question answering task to assess the generalizability of the instruction-tuned model. Using six commonsense reasoning benchmarks, we compared the EntGPT-I model against existing supervised and zero-shot approaches. The accuracy of these models is reported in Table~\ref{tab:qa}. We observed that EntGPT-I not only attains SOTA performance on six QA datasets in a zero-shot setting but also achieves comparable, and sometimes superior, performance compared to supervised methods. 

\subsection{Ablation Study}
\begin{table}[t]
\centering
\footnotesize
\caption{Ablation study on 8 datasets with GPT3.5 backbone. The highest scores are in bold. 
}
\begin{tabular}{p{4.7cm}| p{0.9cm}p{1.0cm}p{1.0cm}p{0.8cm} | p{0.6cm}p{0.9cm}p{0.6cm}p{0.8cm} |p{0.8cm}}
\toprule
\textbf{Ablation}
& \textbf{KORE} & \textbf{OKE15} & \textbf{OKE16} &
\textbf{REU} & \textbf{RSS} & \textbf{ACE04} &
\textbf{MSN} & \textbf{AQU} & \textbf{avg} \\
\midrule
\textbf{EntGPT-P} \scriptsize{w/o Aug. (step 2)} & 0.707	&0.696	&0.687	&0.688	&0.767	&0.853	&0.821 &0.753	& 0.747\\
\textbf{EntGPT-P} \scriptsize{ w/o BLINK (step 1)} & \textbf{0.722}	& \textbf{0.769}	&0.748	&0.676	&0.794	&0.890	&\textbf{0.878}	&\textbf{0.865}	& 0.793\\
\textbf{EntGPT-P} & 0.716	&0.767 & \textbf{0.770}	& \textbf{0.785} & \textbf{0.808}	& \textbf{0.918}	& 0.867 &0.791	& \textbf{0.803} \\

\midrule

\textbf{EntGPT-I} \scriptsize{w/o candidates} & 0.688	& 0.717	& 0.714 &0.739 &0.765	&0.861	&0.832	&0.732	& 0.756\\
\textbf{EntGPT-I} & \textbf{0.753}	&\textbf{0.825} & \textbf{0.819}	& \textbf{0.808} & \textbf{0.825}	& \textbf{0.937}	& \textbf{0.922} &\textbf{0.906}	& \textbf{0.849} \\
\bottomrule
\end{tabular}

\label{tab:ablation_study}
\vspace{-0.2cm}
\end{table}

Here we show the results of an ablation experiment on eight of the ten entity disambiguation benchmarks -- the two very large datasets WIKI and CWEB could not be considered in the ablation study due to budget constraints. This ablation study compared the effectiveness of the entity candidates generation \textit{(Step-1)} on both EntGPT-P and EntGPT-I, as well as the importance of the augmentation by auxiliary content \textit{(Step-2)} on EntGPT-P. Note that without Step-1, EntGPT-I should no longer be considered an instruction-tuned model but rather as a fine-tuned model.

Results of the ablation experiment are shown in Table~\ref{tab:ablation_study} (top). We found that removal of the BLINK entity candidate generation model from the candidates set of EntGPT-P (\textit{i.e.}, use of only the Prior model) did not substantively impact performance. 
We found that removing the auxiliary content (\textit{i.e.} Step-2) severely decreases the performance of EntGPT-P. This shows that auxiliary content enhances the connections between the mention and the target entity. 

To further elucidate the importance of grounding entities to reducing errors, we fine-tuned GPT3.5 on the AIDA dataset without any entity candidates. Table~\ref{tab:ablation_study} (bottom) shows that removal of these entities produced a sharp decline in the mean micro-\textit{$F_1$} indicating that this entity-grounding is directly responsible for the performance increase. 

\subsection{Case Study}

\begin{table}[t]
\centering
\footnotesize
\caption{Examples from the entity disambiguation task. We compared GPT3.5, EntGPT-P and EntGPT-I on the OKE15 dataset. Mentions are underlined in blue.}
\vspace{-.7cm}

\begin{tabularx}{\linewidth}{ @{} p{.52\columnwidth} X @{} } 
    \toprule
    \textbf{Example} &  \textbf{Model Prediction} \\ \hline
    Blackett spent ten years working at the Cavendish Laboratory as an experimental physicist with Ernest Rutherford and in 1923 became a {\color{blue}\underline{\textsf{fellow}}} of King's college, Cambridge. &
    
    \textbf{GPT3.5:} 
    IEEE
    {\color{red} \XSolid } 
 \newline
    \textbf{EntGPT-P:} None {\color{red} \XSolid }  \newline
    \textbf{EntGPT-I:} Fellow {\color{green}\Checkmark} \\
    \hline
    The man spent time at the Royal Aircraft Establishment in {\color{blue}\underline{\textsf{Farnborough}}}, where he made a major contribution to the design of the Mark XIV bomb sight.
    &

    \textbf{GPT3.5:} Royal Aircraft Establishment {\color{red} \XSolid }  \newline
    \textbf{EntGPT-P:} Royal Aircraft Establishment {\color{red} \XSolid } \newline
    \textbf{EntGPT-I:} Farnborough, Hampshire  {\color{green}\Checkmark} \\

    \hline
    Here Lutuli was confirmed in the {\color{blue}\underline{\textsf{Methodist Church}}} and became a lay preacher.  &

    \textbf{GPT3.5:} Methodist Church of Great Britain {\color{red} \XSolid } \newline
    \textbf{EntGPT-P:} United Methodist Church  {\color{green}\Checkmark}\newline
    \textbf{EntGPT-I:} Methodism {\color{red} \XSolid } \\

    \hline
\end{tabularx}

\label{tab:ed_case}
\vspace{-0.2cm}
\end{table}

Three example cases from the OKE15 dataset are compared across three models (GPT3.5, EntGPT-P, and EntGPT-I) on disambiguation task are illustrated in Fig.~\ref{tab:ed_case}. Given the task of linking the mention Farnborough, with context provided, EntGPT-I correctly identified Farnborough, Hampshire. Yet there were also instances where EntGPT-I failed to correctly link the mention to the desired entity. Several other case studies and error analysis are available in the Appendix.

\section{Discussion}

In this work, we described the Entity GPT model including two variants: prompt-based model (EntGPT-P) and instruction-tuned model (EntGPT-I). By linking natural language mentions of concrete entities  generated by large language models to their corresponding entities in some knowledge base, we showed that the EntGPT approach exhibit exceptional ability in entity disambiguation. Furthermore, we also conclude from our extensive experiments that enhancing entity-based knowledge not only helps in entity disambiguation task but also provides better performance in question answering tasks.

This study also motivates us to explore two different follow-up experiments as future work. We first wish to extend the EntGPT framework to perform entity linking (not just entity disambiguation). Compared to entity disambiguation, entity linking not only requires a model to select correct entities but also to extract mentions. Although previous studies~\cite{DeCao-ICLR'21-GENDRE, Xiao-EMNLP'23-Instruction} have explored the application of generative models to entity linking, they do not consistently demonstrate advantages over supervised methods. Second, we wish to dive deeper into how and why entity disambiguation improves question answering performance. These initial observations point to entity-correlation as a crucial factor leading to the better performance, so it may be wise to incorporate entity correlation more deeply into the question answering paradigm.


\bibliographystyle{elsarticle-num}
\bibliography{sample-base}

\clearpage

\appendix

\section{A Case Study on Entity Disambiguation}

In this section, we show the results of a case study of EntGPT's prediction errors to determine the nature of its mistakes. In general, error-cases were of two forms: (1) incorrect ground truth, and (2) predictions were wrong. 

In the first case, we identified cases where the predicted label was, in our opinion, more-correct than the ground truth itself. These cases are indicated a ``GT is incorrect'' in Table~\ref{tab:errors}. We estimate about one-third of the error can be accounted by errors in the ground truth.

\begin{table}[t]
    \centering
    \caption{Comparison of the ground truth labels and the labels predicted by the EntGPT-P model. In many error cases the EntGPT-P model produces labels that, in our opinion, are more accurate than the labeled ground truth. In other cases where the ground truth was indeed correct, the errant prediction was close.}
\scriptsize{
    \begin{tabular}{p{1cm}p{1.5cm}p{4.5cm}p{4.9cm}p{2.3cm}}
\toprule
\textbf{Dataset} & \textbf{Severity} & \textbf{Ground Truth (GT)} & \textbf{Prediction} & \textbf{Source of Error} \\
\midrule
ACE04	&high	&Ministry of Defense (Iran)	&Ministry of Defense (Japan)	&Step 3  \\
ACE04	&low	&President of Egypt	&President	&Step 3 \\
ACE04	&low	&Gaza City	&Gaza Strip	&Step 2 \\
ACE04	&low	&Volvo	&Volvo Cars	&Step 3    \\
KORE	&low	&First Ladies of Argentina	&First Lady	&Step 2 \\
KORE	&high	&Justin Bieber	&Justin I	&Step 2 \\
KORE	&high	&Lady Gaga	&Gwen Stefani	&Step 3 \\
KORE	&high	&Paul Allen	&NULL	&Step 2 \\
AQU	&high	&Cancer	&Lung Cancer	&Step 3 \\
AQU	&high	&Tissue (biology)	&Facial tissue	&Step 3 \\
CWEB	&high	&Head	&Head (company)	&Step 2 \\
CWEB	&none	&Hillsborough County, Florida	&Hillsborough, North Carolina	&GT is incorrect \\
CWEB	&low	&Lake Wylie	&Lake Wylie, South Carolina	&Step 3 \\
CWEB	&none	&Australia Cricket Team	&Australia	&GT is incorrect \\
MSN	&none	&New York City	&New York	&GT is incorrect \\
MSN	&none	&University of Alabama	&Alabama Crimson Tide football	&GT is incorrect \\
MSN	&high	&World Trade Center	&Collapse of the World Trade Center	&Step 3 \\
OKE15	&none	&Fellow	&Research Fellow	&GT is incorrect \\
OKE15	&low	&Cambridge	&University of Cambridge	&Step 2 \\
OKE15	&none	&Principal (academia)	&Head teacher	&GT is incorrect \\
OKE15	&low	&Faculty (academic staff)	&Professor	&Step 2 \\
OKE15	&none	&Officer	&Officer (armed forces)	&GT is incorrect \\
OKE16	&none	&Director (business)	&Executive director	&GT is incorrect \\
OKE16	&none	&Germany	&Nazi Germany	&GT is incorrect \\
OKE16	&none	&Czechs	&Czech Republic	&GT is incorrect \\
OKE16	&none	&Sorbonne	&University of Paris	&GT is incorrect \\
REU	&none	&Georgia Power	&Georgia (U.S. state)	&GT is incorrect \\
REU	&low	&Lloyds Bank of Canada	&Lloyds Bank	&Step 3 \\
RSS	&none	&Steve Jobs	&Apple Inc.	&GT is incorrect \\
RSS	&low	&Pro Bowl	&Super Bowl	&Step 3 \\
RSS	&none	&Eric Kearney	&Cincinnati	&GT is incorrect \\
RSS	&high	&Cleveland Browns	&Cleveland	&Step 3 \\
\bottomrule
\end{tabular}
}
    \label{tab:errors}
\end{table}

In other cases, EntGPT-P did make mistakes. However, we found that in many of these cases these mistakes were still reasonable predictions as these predictions partially matched the ground truth. For example, in the case where the ground truth is the President of Egypt (line 2 in Table~\ref{tab:errors}), the prediction of ``President'' was reasonable but too broad.  We further analyzed the source of these errors and found that most of the mistakes were caused by the multiple choice selection step (Step 3). The augmentation step (Step 2) occasionally failed to provide useful auxiliary content.


\clearpage

\begin{table}[t]
\centering
\scriptsize
\caption{Examples from two QA tasks: ARC-C (left), and OBQA (right), showing the performance of different language models on a particular question. Here, GPT3.5 refers to the foundational GPT3.5 model, and LLAMA2 refers to the Llama2-70B-chat.}
\begin{tabular}{c|p{.7\columnwidth}|p{3.2cm}}

    \toprule

    \textbf{} & \textbf{Example} &  \textbf{Model Prediction} \\
    \hline

   \multirow{15}{*}{\rotatebox[origin=c]{90}{ARC-C}} & An astronaut drops a 1.0 kg object and a 5.0 kg object on the Moon. Both objects fall a total distance of 2.0 m vertically. Which of the following best describes the objects after they have fallen a distance of 1.0 m. 

    (A) They have each lost kinetic energy.\newline
    (B) They have each gained the same amount of potential energy.\newline
    (C) They have each lost the same amount of potential energy.\newline
    {\color{blue}(D) They have each gained one-half of their maximum kinetic
    energy} &

    \multirowcell{7}[0pt][l]{GPT3.5: B {\color{red} \XSolid } \\
    EntGPT-I: C {\color{red} \XSolid }  \\
    LLAMA2: A {\color{red} \XSolid }  } 

    \\
    \cmidrule(lr{2em}){2-3}

    &

     In the equatorial Pacific Ocean, upwelling moves cold, deep sea water to the surface. This water is rich in nutrients and dissolved carbon dioxide gas. During an El Niño event, upwelling declines. This causes surface waters to rise in temperature. Which of the following is most likely to occur in the equatorial Pacific Ocean during an El Niño year? 
    
    {\color{blue}(A) decrease in release of carbon dioxide to the atmosphere }\newline
    (B) increase in photosynthesis by algae in surface waters \newline
    (C) increase in dissolved oxygen in surface waters \newline
    (D) decrease in precipitation over ocean water &

    \multirowcell{9}[0pt][l]{GPT3.5: B {\color{red} \XSolid } \\
    EntGPT-I: A {\color{green} \Checkmark } \\
    LLAMA2: B {\color{red} \XSolid }  } 

    \\
    \midrule
    
    \multirow{12}{*}{\rotatebox[origin=c]{90}{OBQA}} &
     
    
    Predators eat:\newline
    (A) lions 
    (B) humans 
    {\color{blue}(C) bunnies} (D) grass \newline &

        \multirowcell{3}[0pt][l]{GPT3.5: B {\color{red} \XSolid } \\
    EntGPT-I: C {\color{green} \Checkmark }\\
    LLAMA2: A {\color{red} \XSolid } } 
    \\
    \cmidrule(lr{2em}){2-3}

      & The summer solstice in the northern hemisphere is four months before: \newline
    (A) May (B) July (C) April {\color{blue}(D) October} \newline &

            \multirowcell{3}[0pt][l]{GPT3.5: B {\color{red} \XSolid }\\
    EntGPT-I: B {\color{red} \XSolid } \\
    LLAMA2: C {\color{red} \XSolid }} 
    \\
    \cmidrule(lr{2em}){2-3}

     & They studied the soil by using: \newline
    (A) plants (B) a telescope (C) roots {\color{blue}(D) a microscope} \newline &

    \multirowcell{3}[0pt][l]{GPT3.5: D {\color{green} \Checkmark }\\
    EntGPT-I: C {\color{red} \XSolid }\\
    LLAMA2: D {\color{green} \Checkmark }} 
    \\
   \cmidrule(lr{2em}){2-3}


      & If you were attacked by a shark and had to punch it sharply where it pulls in air from, you’d use your hand to make contact with:
    
    (A) its snout {\color{blue}(B) its gills} (C) its nose (D) its belly &

        \multirowcell{3}[0pt][l]{GPT3.5: A {\color{red} \XSolid }\\
    EntGPT-I: B {\color{green} \Checkmark } \\
    LLAMA2: A {\color{red} \XSolid }} 
    \\
    \bottomrule


\end{tabular}
\label{tab:qanew}
\vspace{-0.2cm}
\end{table}

\section{Case Study on Question Answering}

To better understand the results entity linking provides to LLMs, we also conducted a brief error analysis on two of the QA datasets: ARC-C and OBQA. ARC-C is comprised of science exam questions collected from various sources; OBQA is a collection of multiple-choice elementary-level science questions. As reported in Table~\ref{tab:qa}, we performed our experiments with three different models: Llama2-chat, GPT3.5, and EntGPT-I with a GPT3.5 foundation model. We highlight some of the interesting cases as observed in Fig.~\ref{tab:qanew} to assess how different models behave under varying conditions. 

Although we do not delve into a comprehensive examination of errors and their various types, we use these specific instances to provide some analysis and context. We found that all of the models fail on mathematical reasoning: \textit{e.g.}, ``An astronaut drops a 1.0kg object$\ldots$'', which requires an understanding of kinetic energy and the use of some logical deduction. 
We therefore conclude that enhancing the entity-based knowledge does not necessarily help the LLMs in mathematical reasoning.


Moreover, we performed an analysis on cases where the questions required both commonsense and logical reasoning. We found that entity-based knowledge provided the models with a better understanding of the context and therefore helps to mitigate hallucination, with some exceptions. As highlighted in Fig~\ref{tab:qanew}, the examples that require both commonsense knowledge and logical deduction such as the predators-prey relationship and the shark example do benefit from grounding the entities. However, we did sometimes encounter instances when entity linking did not improve performance. For example, in the soil case, the entity-linking lead the model astray, finding ``roots'' rather than the more-direct correct answer ``microscope''. Although these errors are unfortunate, they still leave the door open for additional improvement in future work.

\end{document}